\documentclass[journal,twoside,web]{ieeecolor}
\usepackage{generic}
\usepackage{cite}
\usepackage{amsmath,amssymb,amsfonts}
\usepackage{algorithmic}
\usepackage{graphicx}
\usepackage{algorithm,algorithmic}
\usepackage{textcomp}
\usepackage{CJKutf8}
\usepackage{booktabs}
\usepackage{multirow}
\usepackage{multicol}
\usepackage{longtable}
\usepackage{makecell}
\usepackage[misc]{ifsym}
\usepackage{xcolor, colortbl}
\usepackage{hyperref}
\hypersetup{hidelinks}

\def\BibTeX{{\rm B\kern-.05em{\sc i\kern-.025em b}\kern-.08em
    T\kern-.1667em\lower.7ex\hbox{E}\kern-.125emX}}
\begin{document}
\begin{CJK}{UTF8}{gbsn}
\definecolor{mygray}{gray}{.9}
\definecolor{mypink}{rgb}{.99,.91,.95}
\definecolor{mygreen}{rgb}{.9,.99,.9}

\title{DongYuan: An LLM-Based Framework for Integrative Chinese and Western Medicine Spleen-Stomach Disorders Diagnosis}

\author{
Hua Li\textsuperscript{†}, Yingying Li\textsuperscript{†}, Xiaobin Feng\textsuperscript{†}, Xinyi Fu,
Lifeng Dong, Qingfeng Yang, Yanzhe Chen, Xiaoju Feng, Zhidong Cao, Jianbin Guo\textsuperscript{*}, Yanru Du\textsuperscript{*}
\thanks{Yingying Li, Xiaobin Feng and Lifeng Dong are with Beijing Wenge Technology Co., Ltd.,
Beijing, China (e-mail: \{yingying.li, xiaobin.feng, lifeng.dong\}@wenge.com).}
\thanks{Hua Li, Qingfeng Yang, Yanzhe Chen, Xiaoju Feng and Yanru Du are with
Hebei Provincial Hospital of Traditional Chinese Medicine, Shijiazhuang, China
(e-mail: 1430885043@qq.com, 578511160@qq.com, chenyanzhe718@163.com, 254038089@qq.com, zyydyr@163.com).}
\thanks{Xinyi Fu, Lifeng Dong, Zhidong Cao and Jianbin Guo are with
Institute of Automation, Chinese Academy of Sciences, Beijing, China
(e-mail: fuxinyi2026@ia.ac.cn, jianbin.guo@ia.ac.cn, zhidong.cao@ia.ac.cn).}
\thanks{Jianbin Guo is also with Tianjin University, Tianjin, China.}
\thanks{\textsuperscript{†}Yingying Li, Xiaobin Feng and Hua Li contributed equally to this work.}
\thanks{\textsuperscript{*}Jianbin Guo and Yanru Du are the corresponding authors.}
}
\maketitle

\begin{abstract}
The clinical burden of spleen-stomach disorders is substantial. While large language models (LLMs) offer new potential for medical applications, they face three major challenges in the context of integrative Chinese and Western medicine (ICWM): a lack of high-quality data, the absence of models capable of effectively integrating the reasoning logic of traditional Chinese medicine (TCM) syndrome differentiation with that of Western medical (WM) disease diagnosis, and the shortage of a standardized evaluation benchmark. To address these interrelated challenges, we propose DongYuan\footnote{\textit{DongYuan} is named after \textit{Li Dongyuan} (李东垣), the honorific title of \textit{Li Gao} (李杲), the founder of the \textit{spleen-stomach theory} in TCM, whose academic legacy is deeply integrated into the methodological foundation of this work.}, an ICWM spleen-stomach diagnostic framework. Specifically, three ICWM datasets (SSDF-Syndrome\footnote{"SSDF" stands for "spleen-stomach diagnostic framework", which underpins the naming convention for all core components of the proposed \textit{DongYuan} framework.}, SSDF-Dialogue, and SSDF-PD) were curated to fill the gap in high-quality data for spleen-stomach disorders. We then developed SSDF-Core, a core diagnostic LLM that acquires robust ICWM reasoning capabilities through a two-stage training regimen of supervised fine-tuning (SFT) and direct preference optimization (DPO), and complemented it with SSDF-Navigator, a pluggable consultation navigation model designed to optimize clinical inquiry strategies. Additionally, we established SSDF-Bench, a comprehensive evaluation benchmark focused on ICWM diagnosis of spleen-stomach disorders. Experimental results demonstrate that SSDF-Core significantly outperforms 12 mainstream baselines on SSDF-Bench. DongYuan lays a solid methodological foundation and provides practical technical references for the future development of intelligent ICWM diagnostic systems.
\end{abstract}


\section{Introduction}

Spleen-stomach disorders are among the most common, highly recurrent, and challenging conditions in integrative Chinese and Western medicine (ICWM) clinical practice, and their management efficacy is closely linked to overall population health. Contemporary global health research has identified a growing burden of gastrointestinal (GI) diseases: a recent U.S. study \cite{peery2025burden} reported 315,065 newly diagnosed GI cancer cases and 281,413 GI disease-related deaths, underscoring the heavy pressure of this disease spectrum on public health systems. Chronic and recurrent functional GI disorders are also escalating worldwide: in 2021, gastroesophageal reflux disease affected approximately 825.6 million people, with its incidence, prevalence, and years lived with disability (YLDs) rising sharply since 1990 and projected to continue increasing over the next decade, especially within the working-age population \cite{mo2025global}. Regional analyses further indicate that gastric cancer remains a leading cause of cancer-related deaths in China, and GI malignancies account for a considerable share of global cancer incidence and mortality \cite{han2024cancer}. Collectively, these epidemiological trends present sustained challenges to the clinical management of spleen-stomach disorders.

Traditional Chinese medicine (TCM) adopts a holistic framework and prioritizes syndrome differentiation-based treatment, yet its heavy reliance on physicians’ personal experience can introduce subjective bias, resulting in inconsistent diagnostic conclusions and therapeutic strategies for identical clinical presentations across different practitioners \cite{lam2019icd11}. In contrast, Western medicine (WM) excels in objective diagnosis by precisely identifying structural lesions via endoscopy and histopathology to enable targeted interventions. However, it is less adaptable to individual heterogeneity and dynamic disease progression, and often overlooks systemic functional abnormalities—especially for functional disorders lacking distinct organic lesions \cite{liu2024traditional}. Therefore, the organic integration of TCM syndrome differentiation with WM objective diagnosis represents a promising approach to enhance diagnostic accuracy, promote early intervention, and support personalized and adaptive clinical management for such conditions \cite{jin2023research}.

Computer-aided diagnosis (CAD) \cite{CAD2024review} has become indispensable in managing spleen-stomach disorders, evolving from rule-driven to data-driven paradigms. Early rule-based expert systems and data mining techniques usually distilled clinical experience into diagnostic rules, yet struggled to adapt to dynamic real-world symptom patterns \cite{yang2022study}. Subsequent machine learning \cite{lin2025development} and deep learning \cite{zhang2021deep} approaches further advanced the field, such as cross-modal attention for non-invasive assessment \cite{qiao2024intelligent} and knowledge graph-based methods for structured knowledge representation  \cite{KG4SSD2022}. Despite these progresses, existing methods suffer from limited representational capacity and insufficient reasoning depth, making them inadequate for the complex clinical reasoning required in ICWM diagnosis of spleen-stomach disorders.

The rapid advancements in large language models (LLMs) have opened new avenues for addressing existing limitations \cite{zijia2024application,waisberg2023gpt4,singhal2025toward}. In this field, a synergistic "data–model–benchmark" development paradigm has gradually taken shape. In general medicine, notable advances include the MultiMedQA benchmark and Med-PaLM models \cite{singhal2023large}, the Baichuan-M2 model equipped with validation and reinforcement learning mechanisms \cite{Dou2025BaichuanM2SM}, the multilingual MMedC corpus, MMed model series, and MMedBench \cite{qiu2024towards}, as well as the MCC model with multi-model collaborative reasoning \cite{sun2026model}. Alongside these advances, TCM-specific LLMs have also been developed to accommodate the unique nature of TCM clinical practice \cite{TCM2025LLM}, including the Lingdan model series supported by specialized datasets \cite{hua2024lingdan}, ZhongJingGPT with optimized consultation logic \cite{zhongJing2025}, the TCM-DS model focused on dietary therapy \cite{li2025tcmds}, the Qibo model and benchmark \cite{QiboA2025}, and the multimodal BenCao assistant \cite{Xie2025BenCaoAI}.

Despite significant advancements in applying LLMs to medicine, substantial gaps remain:

\textbf{At the data level}, high-quality diagnostic ICWM datasets for spleen-stomach disorders are scarce. \textbf{First}, significant disparities exist between the terminological systems of TCM and WM, with no unified standard for mapping TCM syndromes to WM diseases—hindering the mutual integration and coordinated utilization of TCM syndrome differentiation logic and WM examination indicators. \textbf{Second}, existing datasets predominantly follow a simplistic input-output mapping paradigm (e.g., symptoms to syndromes), lacking explicit annotation of expert clinical reasoning processes. \textbf{Third}, most available data are dominated by single-turn question-and-answer exchanges, with few large-scale, standardized multi-turn proactive consultation datasets specialized for spleen-stomach disorders.

\textbf{At the model level}, the performance of existing medical LLMs remains limited in complex, domain-specific disease scenarios \cite{ullah2024challenges}. \textbf{First}, most models adopt a static question-answering paradigm and lack strategic, proactive follow-up inquiry capabilities essential for clinical consultation tasks. Existing active inquiry methods are primarily designed for WM scenarios \cite{Wang2025EmpoweringMM,Yu2025FromPT}. In particular, inquiry approaches relying solely on LLMs exhibit dispersed, hard-to-converge strategies due to the models’ inherent generative nature, resulting in low coverage of clinically critical symptoms and poor consultation efficiency. \textbf{Second}, current medical LLM development relies mainly on supervised fine-tuning (SFT). While SFT allows models to follow task-specific instructions, it only captures surface-level patterns without encoding expert preferences. Thus, SFT-only models often lack the nuanced clinical judgment needed for real-world decision-making. \textbf{Moreover}, most models lack explicit reasoning constraints for ICWM diagnosis, failing to embody the integrated clinical logic of TCM syndrome differentiation combined with WM examinations.

\textbf{At the evaluation level}, a systematic evaluation framework for ICWM-based spleen-stomach disorder diagnosis has not yet been established. \textbf{First}, most current evaluations focus on general medical tasks and single medical systems (pure TCM or pure WM), failing to adapt to the disease-specific ICWM diagnosis. \textbf{Second}, evaluation mechanisms are predominantly result-oriented \cite{long2025large}, focusing solely on accuracy metrics and neglecting key dimensions like the interpretability of TCM syndrome differentiation reasoning. This makes it difficult to comprehensively assess a model’s clinical practicality. \textbf{Additionally}, some existing benchmarks suffer from incomplete data disclosure, compromising the reproducibility and comparability of evaluation results.

\begin{figure*}[htbp]
\centering
\includegraphics[width=1\textwidth]{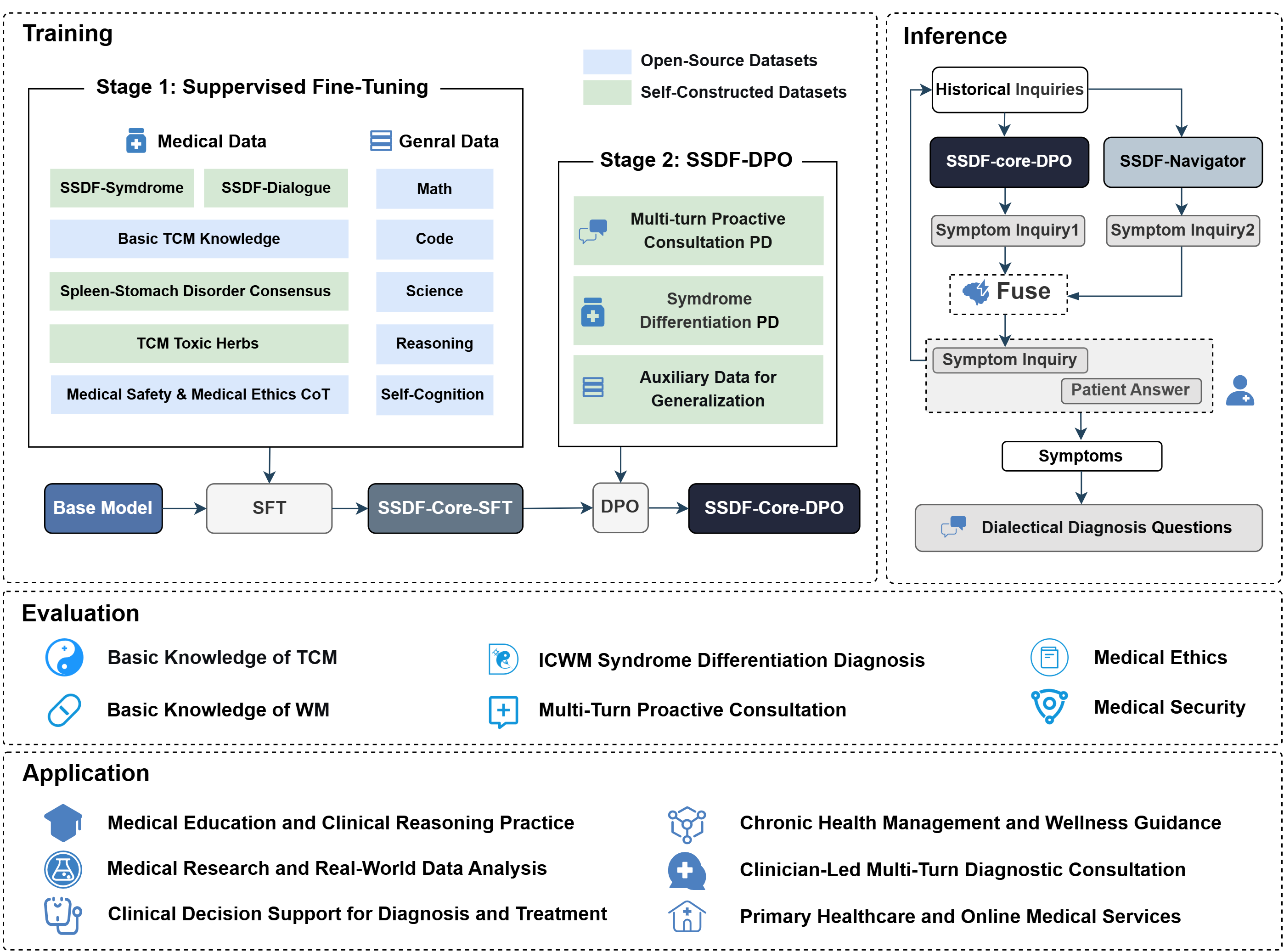}
\caption{Overview of DongYuan. The framework integrates three specialized datasets (SSDF‑Syndrome, SSDF‑Dialogue, SSDF‑PD), a core diagnostic LLM trained via a two‑stage pipeline (SFT → DPO), a pluggable consultation navigation model (SSDF‑Navigator), and a comprehensive evaluation benchmark (SSDF‑Bench). Application scenarios cover a diverse array of ICWM clinical and research-oriented tasks.}
\label{fig:overview}
\end{figure*}

To address these challenges, we propose DongYuan, a comprehensive framework focused on ICWM-based intelligent diagnosis of spleen-stomach disorders, covering specialized datasets, models, and an evaluation benchmark. The overall framework is shown in Fig.\ref{fig:overview}. First, we constructed three datasets: \textbf{SSDF-Syndrome} (for syndrome differentiation), \textbf{SSDF-Dialogue} (for multi-turn proactive consultation), and \textbf{SSDF-PD} (for direct preference optimization). Domain experts’ clinical reasoning knowledge is explicitly encoded in a structured format within all three datasets, and SSDF-Dialogue supports multi-turn proactive consultations with standardized inquiry strategies. This suite of datasets systematically bridges the gap in high-quality domain-specific data. Second, we developed a core diagnostic LLM (\textbf{SSDF-Core}) alongside a pluggable consultation navigation model (\textbf{SSDF-Navigator}). Built on a general-purpose language model, SSDF-Core adopts a two-stage training pipeline with SFT and preference optimization (DPO), which balances general medical capabilities and ICWM diagnostic expertise. SSDF-Navigator employs a hybrid training strategy that integrates offline reinforcement learning with behavior cloning. It is designed to learn clinically consistent inquiry strategies that mitigate the divergent inquiry behavior inherent in the core diagnostic LLM during multi-turn proactive consultation. This collaboration improves diagnostic efficiency and critical symptom coverage. Finally, we established \textbf{SSDF-Bench}. Beyond result-oriented accuracy metrics, it introduces chain-of-thought indicators to quantify clinical reasoning integrity and alignment with real diagnostic logic from a process perspective. Experimental results demonstrate that SSDF-Core outperforms mainstream LLMs on SSDF-Bench. All datasets, model implementations, and the benchmark are publicly available\footnote{\href{https://github.com/DYJG-research/TCWMSDT}{https://github.com/DYJG-research/TCWMSDT}}. The main contributions of this study are summarized as follows:
\begin{itemize}
    \item \textbf{Data}: We release the first three open-source specialized datasets for ICWM spleen-stomach disorders diagnosis, which are sourced from real-world clinical records and rigorously annotated by ICWM physicians.
    \item \textbf{Model}: We proposed SSDF-Core and SSDF-Navigator. The two-stage training pipeline of SSDF-Core delivers effective ICWM diagnostic performance without compromising its general medical capabilities. SSDF-Navigator optimizes consultation strategies via hybrid training and guides SSDF-Core in multi-turn proactive consultation, improving both efficiency and interpretability.
    \item \textbf{Benchmark}: We establish SSDF-Bench, an open-source benchmark focused on ICWM spleen-stomach diagnosis using real-world post-recovery clinical data, offering a standardized tool for domain-specific model evaluation.
\end{itemize}

\section{Methods and Materials}
\subsection{Datasets Construction}
To address the lack of specialized ICWM datasets for spleen-stomach disorder diagnosis, we curated three datasets. The raw data, sourced from real-world retrospective clinical records of spleen-stomach disorders, were approved by the Ethics Committee of Hebei Provincial Hospital of Traditional Chinese Medicine (Approval No.: AF/SC-08/03.0), guaranteeing adherence to all ethical and regulatory requirements during data collection, processing, and utilization. The use of real-world post-recovery records ensures that the data reflects actual diagnostic reasoning and treatment practices, as well as clinically validated therapeutic pathways, providing a solid foundation for modeling effective diagnostic patterns. The original medical records contained 14 core fields, including the clinical reasoning for TCM syndrome differentiation and the ICWM treatment plan, providing a comprehensive information foundation for subsequent data construction. 

For all three datasets, data preprocessing followed a standardized protocol: de-identification to remove personally identifiable information; data cleaning to eliminate duplicate records and non-textual content while standardizing professional terminology; and data filtering to retain high-quality samples with complete consultation records, clear syndrome differentiation conclusions, as well as comprehensive TCM and WM diagnostic information. This pipeline ensures data privacy, terminological consistency, and information integrity, providing a reliable foundation for training robust diagnostic models.

To ensure annotation consistency, a dual mechanism of double-blind annotation and third-party arbitration was implemented. The Kappa coefficient between the two annotating ICWM physicians was 0.86, with a disagreement rate below 12\%. Disputed samples were adjudicated by a senior chief physician from the partner hospital, ensuring annotation reliability and minimizing subjective bias.

To prevent data leakage and ensure the model generalizes to unseen cases, we enforce strict non-overlap at two levels:  first, the original cases used to construct SSDF-Syndrome and SSDF-Dialogue (for SFT) are disjoint from those used for SSDF-PD (for DPO). Second, all evaluation samples in SSDF-Bench are drawn exclusively from held-out cases that do not appear in any of the training sets, thereby guaranteeing unbiased assessment.

\begin{figure}[t]
\centering
\includegraphics[width=0.5\textwidth]{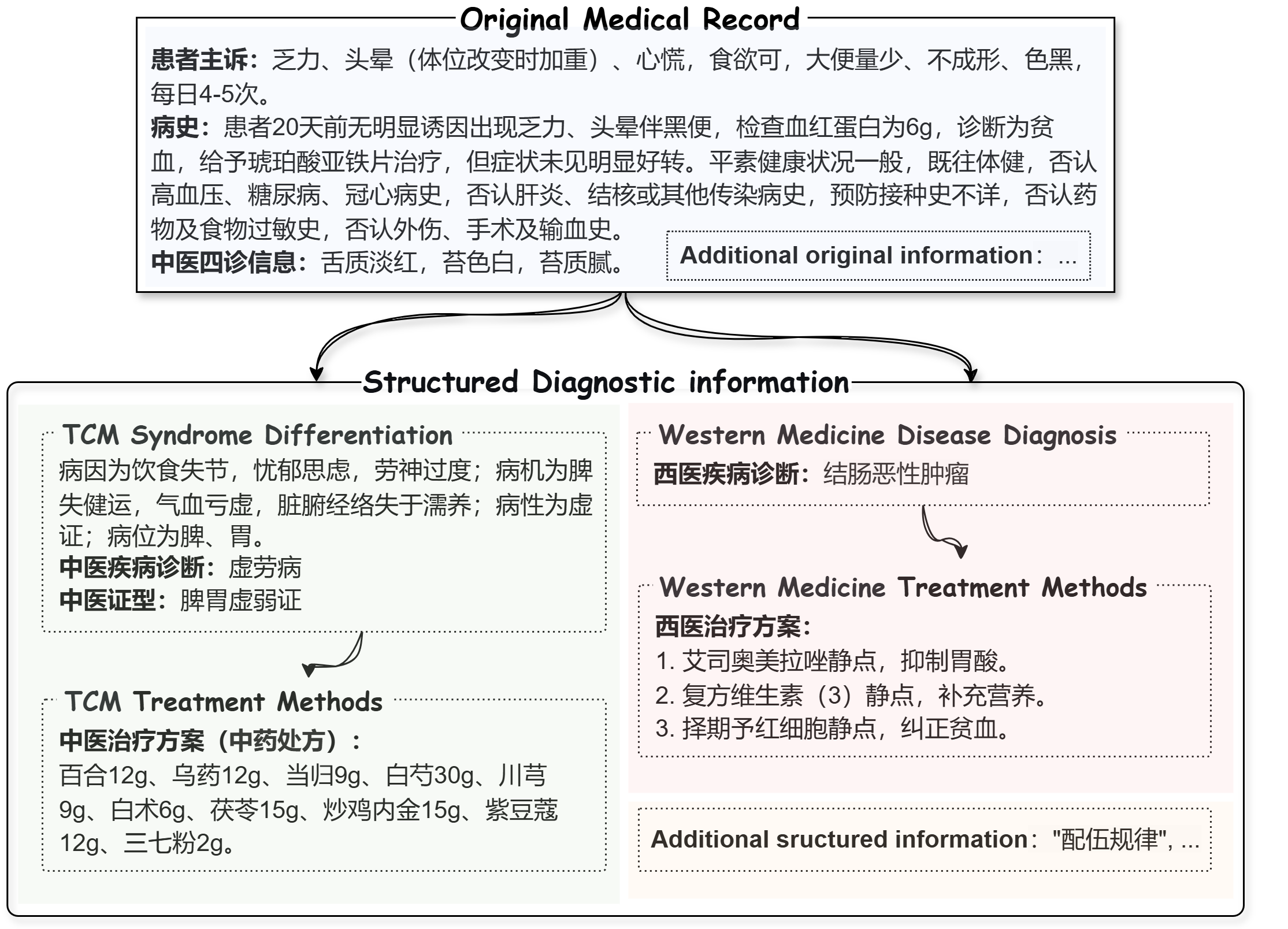}
\caption{\small An example from the SSDF-Syndrome dataset. Each sample is constructed from original medical records, presenting fully structured diagnostic information that explicitly integrates TCM syndrome differentiation, WM disease diagnosis, and other structured clinical information.}
\label{fig:SSDF-Syndrome}
\end{figure}

\subsubsection{SSDF-Syndrome}
SSDF-Syndrome is designed to support ICWM syndrome differentiation for spleen-stomach disorders. It is constructed from patients’ chief complaints and related clinical information, with structured diagnostic information as its final form. Fig. \ref{fig:SSDF-Syndrome} presents a representative example from SSDF-Syndrome.

The construction of SSDF-Syndrome followed a four-step process: 1) Qwen3-32B \cite{yang2025qwen3} was used to extract and summarize key information from the patient's chief complaint in medical case records; 2) the two annotating ICWM physicians verified the accuracy of the extracted key information and supplemented any missing content; 3) the modified key information was re-input into Qwen3-32B to generate a structured clinical result containing seven core fields; 4) based on the structured clinical result, the two annotating ICWM physicians and the senior chief physician collaboratively constructed a clinically coherent diagnostic reasoning process, which together with the structured result constituted the final sample.

Following these steps, a total of 984 clinically plausible samples with explicit diagnostic reasoning that correlates TCM syndrome patterns with WM evidence were constructed, providing a structured resource essential for training models to perform grounded, dual-perspective diagnoses.

\subsubsection{SSDF-Dialogue}
\begin{figure}[t]
\centering
\includegraphics[width=0.4\textwidth]{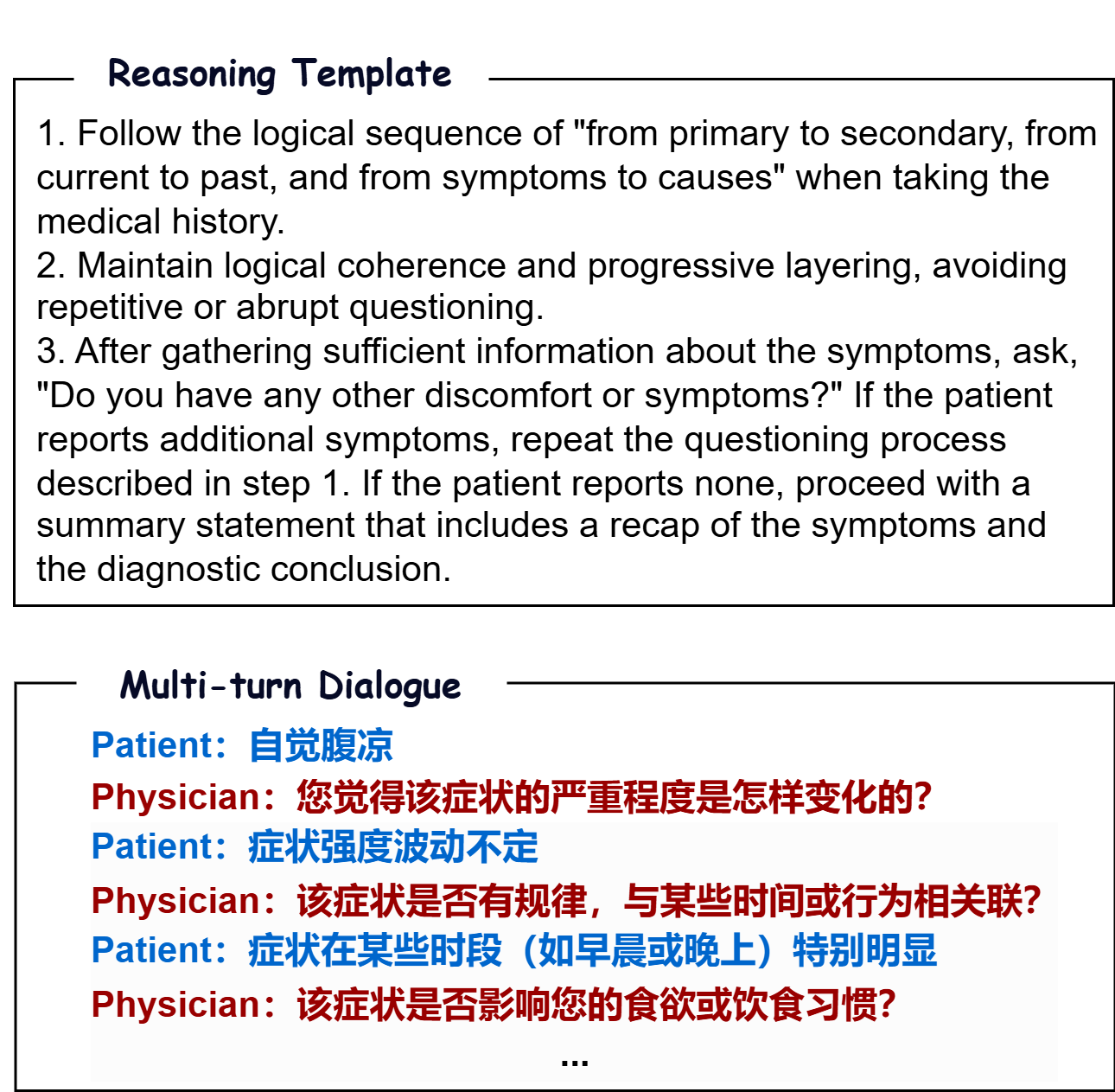}
\caption{\small An example from the SSDF‑Dialogue dataset. The dialogue follows a clinically coherent progression: starting from an initial symptom, the model sequentially inquires about severity fluctuation, temporal patterns, and associated impacts on daily activities.}
\label{fig:SSDF-Dialogue}
\end{figure}

SSDF-Dialogue is built centered on authentic physician-patient conversations from original medical records. Each sample comprises a multi-turn dialogue and a corresponding reasoning template that documents the clinical logic behind each inquiry. Fig \ref{fig:SSDF-Dialogue} presents a representative example from SSDF‑Dialogue.

The reasoning template was developed via an expert-led procedure: 1) the three aforementioned ICWM physicians designed a basic reasoning template structured into an explicit instruction schema, which incorporate historical consultation information, candidate next-step questions, and ICWM-oriented analytical constraints; 2) an information verification module was added into the basic reasoning template for the penultimate inquiry round. This module covers core symptom confirmation, retrospective review of inquiry history, identification of predisposing factors, screening for accompanying symptoms, and closed-loop verification of the ICWM diagnosis; 3) the reasoning template for the last inquiry aggregates the complete historical consultation information to generate structured ICWM diagnostic conclusions.

Following the above construction process, we integrated the reasoning template with the corresponding multi-turn dialogue from the preprocessed original medical cases to form the initial samples. All initial samples subsequently underwent a quality review conducted by the senior ICWM chief physician, who verified the clinical rationality of each sample’s consultation logic and the accuracy of ICWM reasoning.

The final SSDF-Dialogue contains 715 samples with approximately 17,000 dialogue turns, covering 10 common TCM syndrome patterns and achieving a 92\% core symptom coverage rate. Unlike traditional dialogue datasets \cite{zeng2020meddialog}, it explicitly models proactivity of clinical reasoning via a structured reasoning template, which guarantees comprehensive information collection and logically consistent consultations.

\subsubsection{SSDF-PD}
To facilitate effective preference optimization, we constructed SSDF-PD, a dataset containing 12,800 pairwise preference samples for two core tasks: multi-turn proactive consultation and syndrome differentiation.

We established a three-tier preference judgment criterion for sample construction: \textbf{1) Safety First}: responses involving contraindications, obvious syndrome misdiagnoses, or potential clinical risks are directly labeled as rejected; \textbf{2) Syndrome Differentiation Integration}: responses that reasonably integrate WM examination results with TCM syndrome differentiation analysis and follow efficient inquiry pathways are labeled as positive; \textbf{3) Practicality of the Plan}: responses with generally correct diagnostic logic but lacking clinical specificity or operability are labeled as weak responses. These criteria ensure the objectivity and consistency of preference annotations. 

Based on the aforementioned criteria, pairwise preference samples were constructed as follows: 1) for each original clinical case, high-quality outputs generated by our SFT LLM (SSDF-Core-SFT) were selected as positive (chosen) responses after expert verification; 2) corresponding negative (rejected) responses were then constructed by experts who manually injected typical errors into the positive responses, such as logical leaps or omission of key diagnostic evidence. Collectively, these constructed pairwise samples provide a robust foundation for the DPO stage, steering the model toward safe, clinically nuanced outputs aligned with ICWM reasoning.

\subsection{The core diagnostic LLM}
We developed the core diagnostic LLM (SSDF-Core) using a two-stage progressive training paradigm comprising Supervised Fine-Tuning (SFT) and Direct Preference Optimization (DPO), with Qwen3-14B \cite{yang2025qwen3} as the base model. Functionally, this training pipeline is inspired by the hierarchical concept of “skill formation and decision optimization” from the ACT-R cognitive architecture \cite{ACTR2004}. The SFT stage transforms the base model's general knowledge into executable clinical task capabilities through instruction-response alignment \cite{SFT2025}. Subsequently, the DPO \cite{dpo2023} stage incorporates the decision-making preferences of clinical experts to perform fine-grained calibration of the model's outputs, thereby enhancing its professionalism and clinical usability. Through this phased optimization, the model transitions smoothly from general medical cognition to specialized clinical scenarios involving spleen-stomach disorders.

\subsubsection{Supervised Fine-Tuning}
To equip the base model with task-specific capabilities, we constructed a high-quality SFT dataset with a total of 4,000 samples, covering general data and medical data. Among them, medical data serves as the core of the dataset, containing the fully self-constructed SSDF-Syndrome and SSDF-Dialogue. The data statistics for SFT are shown in Table \ref{tab:sft_data_stats}.

\begin{table*}[htbp]
\centering
\caption{Data Statistics of the SFT Dataset}
\renewcommand{\arraystretch}{1.2}
\setlength{\tabcolsep}{6pt}
\begin{tabular}{llcc}
\toprule
\textbf{Type} & \textbf{Name} & \textbf{\# of Instances} & \textbf{Size} \\
\midrule
\multirow{1}{*}{\textbf{General Data}} & Multiple-Choice Questions of Basic Knowledge & 1800 & 9.8 M \\
\midrule
\multirow{6}{*}{\textbf{Medical Data}} & SSDF-Syndrome & 984 & 12.2 M \\
                    & SSDF-Dialogue & 600 & 4.9 M \\
                    & TCM Toxic Herbs (19 Incompatibilities \& 18 Antagonisms) & 419 & 1.4 M \\
                    & Basic TCM Knowledge\cite{hwtcm2024} & 18015 & 90.5 M \\
                    & Spleen-Stomach Disorder Consensus Data & 608 & 2.2 M \\
                    & Medical Safety \& Medical Ethics CoT\cite{ding2025medbenchv4robustscalable} & 1800 & 13.6 M \\
\bottomrule
\end{tabular}
\label{tab:sft_data_stats}
\end{table*}

We performed full-parameter fine-tuning on the base model for 3 epochs, with an initial learning rate of 5e-6 using the AdamW optimizer \cite{adamw2017} and a linear decay strategy \cite{Defazio2023OptimalLD}. Training was conducted on 8 A100 80GB GPUs, achieving an effective batch size of 64 through gradient accumulation with a step size of 8. To optimize training efficiency, we employed bfloat16 mixed precision \cite{kalamkar2019study}, gradient checkpointing \cite{gradientCheckpointing2024}, and Flash Attention \cite{flashAttention2024} acceleration techniques.

The optimization objective was designed to maximize the likelihood of the model generating responses that align with the expert-annotated standard responses in the SFT dataset, defined as:
\begin{align}
    \mathcal{L}_{\text{SFT}}(\theta) = -\frac{1}{N} \sum_{i=1}^{N} \sum_{t=1}^{T_{i}} \log P_{\theta}(y_{t}^{(i)} \mid \mathbf{x}^{(i)}, y_{<t}^{(i)})
\end{align}
where  \( \mathbf{x}^{(i)} \)  represents the  \( i \) -th instruction and its context, and  \( y_{t}^{(i)} \)  is the   \( t \) -th token in the desired response.

The resultant model, SSDF-Core-SFT, acquires task-specific response capabilities, establishing the necessary condition for subsequent preference optimization.

\subsubsection{Direct Preference Optimization}
We further optimized SSDF-Core-SFT via DPO using SSDF-PD. To preserve previously acquired clinical capabilities while aligning with expert preferences, a portion of the SFT data was incorporated during DPO training. Following the relative preference optimization (RPO) approach \cite{rpo2024}, we set the weight parameter $\beta$ to 0.1 and trained for 2 epochs with a maximum sequence length of 10,240 tokens; all other training configurations were identical to those used in SFT.

The optimization objective employed is the standard DPO loss function, which achieves preference alignment by directly optimizing the difference in the model’s conditional generation probabilities between chosen and rejected responses:
\begin{align}
\mathcal{L}_{\text{DPO}}(\theta) = -\mathbb{E}_{(x, y_w, y_l) \sim \mathcal{D}} 
\Bigl[ \log \sigma \bigl( \beta \cdot \Delta r_\theta(x, y_w, y_l) \bigr) \Bigr]
\end{align}
where \( \Delta r_\theta(x, y_w, y_l) = r_\theta(x, y_w) - r_\theta(x, y_l) \) is the reward margin. Here, \( r_\theta(x, y) = \log \frac{\pi_\theta(y|x)}{\pi_{\text{ref}}(y|x)} \) is the implicit reward (log probability ratio). \( x \) is the input instruction, \( y_w \) and \( y_l \) are the chosen and rejected responses, respectively, \( \pi_{\text{ref}} \) is the reference model (here is SSDF-Core-SFT), \( \pi_\theta \) is the policy model being optimized, and \( \sigma \) is the sigmoid function.

The resultant model, SSDF-Core-DPO, achieves enhanced alignment with clinical expertise in ICWM reasoning and serves as the final inference model in our framework.

\subsection{The consultation navigation model}
To constrain the divergent inquiring behavior of the core LLM during multi-turn consultation, we propose SSDF-Navigator, a lightweight pluggable model that learns clinically consistent inquiry strategies to improve diagnostic efficiency and critical symptom coverage.

\subsubsection{Problem formulation}
We formulated multi-turn consultation as a sequential decision-making task, where the process was represented as an ordered sequence of decisions that begins with the patient's chief complaint and progressively integrates the necessary diagnostic information.

\begin{figure*}[htbp]
\centering
\includegraphics[width=0.8\textwidth]{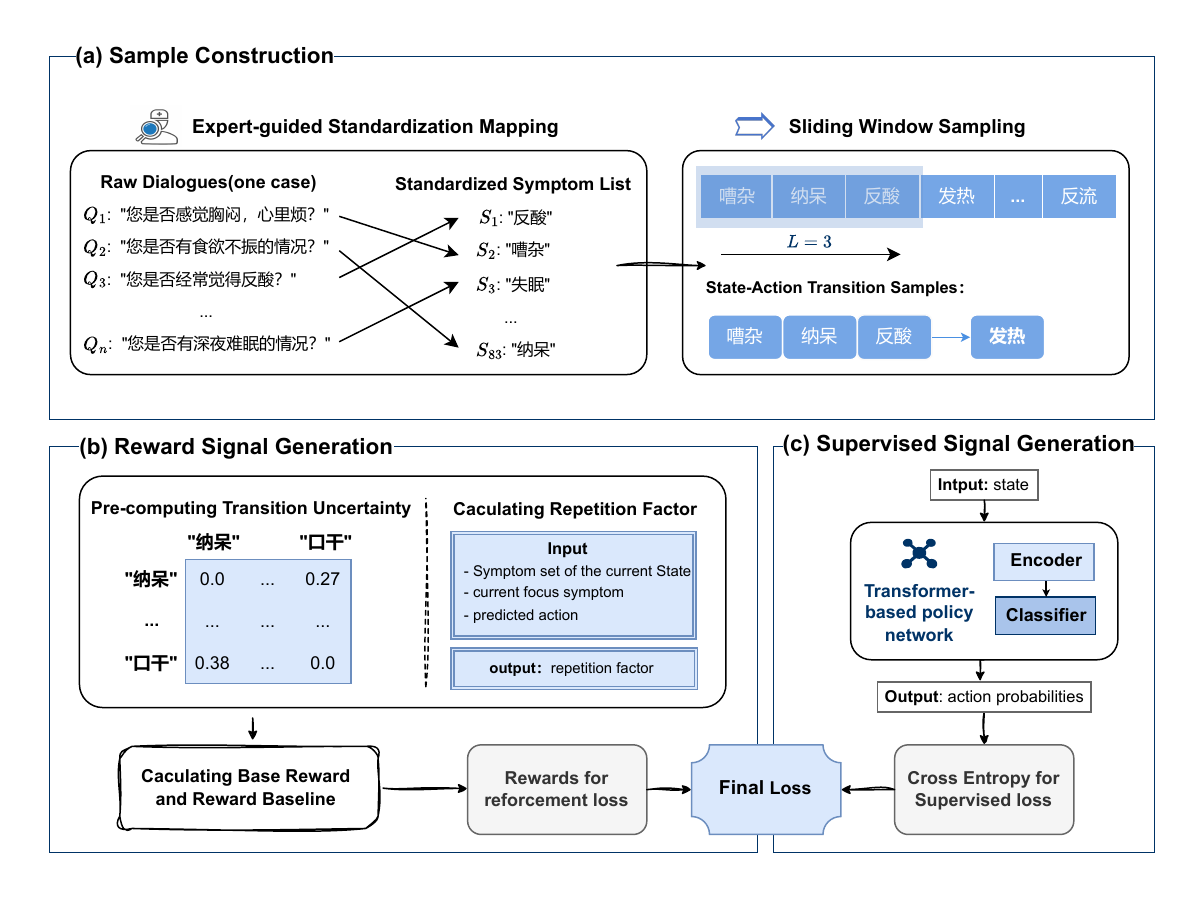}
\vspace{-10pt}
\caption{Overview of SSDF‑Navigator. The model uses a Transformer-based encoder-classifier architecture, with the symptom set from historical dialogue as input, to predict the next symptom. Training combines behavior cloning with offline reinforcement learning. During inference, it collaborates with SSDF‑Core via a candidate-based inquiry selection mechanism to guide multi‑turn proactive consultation.}
\label{fig:SSDF-Navigator}
\end{figure*}

To train SSDF-Navigator, we constructed an offline dataset from authentic physician-patient dialogues. As shown in Fig. \ref{fig:SSDF-Navigator}, under the guidance of the ICWM physicians, original inquiries were mapped to a standardized terminology list comprising 83 core symptoms of spleen-stomach disorders. This modeling approach is based on two clinical assumptions: 1) patient responses mainly confirm or deny symptoms and have a limited impact on subsequent inquiries; 2) standardized symptom categories are more consistent for diagnostic modeling than natural language. Consequently. Thus, each case was represented as a sequence of standardized symptom indices, \( S_i = [s_0, s_1, \dots, s_N] \), where \( s_t \) corresponds to the \( t \)-th inquiry. We then sampled state-action transition pairs $(\text{state}_t, a_t)$ from $S_i$ using a sliding window of length $L$, where $\text{state}_t = [s_{t-L+1}, \dots, s_t]$ is the historical window and $a_t = s_{t+1}$ is the next inquired symptom. The objective is to learn a policy $\pi(a_t \mid \text{state}_t)$ that mimics expert inquiry logic.

\subsubsection{Model architecture and algorithm design}
SSDF-Navigator employed a Transformer-based encoder-classifier architecture to extract contextual representations from historical symptom sequences and predict the next symptom for inquiry. To simultaneously model expert consultation patterns and information-efficient consultation policy, we employ a hybrid training paradigm that combines behavior cloning (BC) with offline reinforcement learning (Offline RL).

The offline RL \cite{offlineRL2020} is based on an information gain-oriented reward function, which is designed to quantify the informational value that a chosen inquiry contributes to the final diagnosis. Guided by the principle that an inquiry significantly reducing the uncertainty of subsequent inquiries is considered to accelerate diagnostic convergence, we compute the reward based on the symptom transition probability distribution in the offline dataset. First, we estimate the conditional probability distribution of symptom transitions with Laplace smoothing:
\begin{align}
P(s_j \mid s_i) = \frac{C(s_i, s_j) + \alpha}{C(s_i) + \alpha \cdot N}
\end{align}
where $C(s_i)$ is the total frequency of $s_i$, $C(s_i,s_j)$ is the frequency of $s_j$ immediately following $s_i$, \( N=83 \) is the total number of standard symptoms, and \( \alpha \) is the smoothing factor. Subsequently, we define the normalized conditional entropy for \( s_i \):
\begin{align}
    H(S \mid s_i) = -\frac{1}{\log N} \sum_{s_j \in \mathcal{S}} P(s_j \mid s_i) \log P(s_j \mid s_i)
\end{align}
where \(H(S \mid s_i)\in[0,1] \), a smaller value of which indicates a more focused and predictable subsequent inquiry path. The base reward is defined as the information gain of the predicted action \( a \):
\begin{align}
    R_{\text{info}}(a) = 1 - H(S \mid a)
\end{align}
To align with clinical practice, we introduce a heuristic repetition factor $\lambda$ to avoid ineffective consecutive repetitions while tolerating revisits to key symptoms. Let $\mathcal{H}$ denotes the set of symptoms in the current state $\text{state}_t$:
\begin{align}
    \lambda(a, \mathcal{H}) =
    \begin{cases}
    0.3, & \text{if } a = x \\
    1.5, & \text{if } a \in \mathcal{H} \ \text{and} \ a \neq x \ \\
    1.0, & \text{otherwise}
    \end{cases}
\end{align}
Finally, the immediate reward \( r_t \) is scaled by a coefficient $K$ to ensure its magnitude is compatible with the supervised loss:
\begin{align}
    r_t = R_{\text{info}}(a) \cdot \lambda(a, \mathcal{H}) \cdot K
\end{align}

The BC generates supervisory signals that guide SSDF-Navigator to learn expert consultation patterns, with cross-entropy loss calculated as:
\begin{align}
\mathcal{L}_{\text{BC}}(\theta) = -\frac{1}{B} \sum_{i=1}^{B} \log \pi_\theta(a^*_t \mid state_t)
\end{align}
where \( \pi_\theta \) is the parameterized policy, \( B \) is the batch size, and $a^*_t$ is the expert action for the state $state_t$.

To integrate BC and RL signals, we designed two mechanisms: \textbf{Reward-Weighted Behavior Cloning} (RWBC) and \textbf{Reward-Added Behavior Cloning} (RABC). RWBC weighted the BC loss by the immediate reward to prioritize information-efficient decisions when imitating experts:
\begin{align}
\mathcal{L}_{\text{RWBC}}(\theta) = -\frac{1}{B} \sum_{i=1}^{B} r_t \cdot \log \pi_\theta(a^*_t \mid state_t)
\end{align}
RABC combines the BC loss with a reward-based policy optimization (PO) loss via linear weighting, with the PO loss defined as:
\begin{align}
\mathcal{L}_{\text{PO}}(\theta) = -\frac{1}{B} \sum_{i=1}^{B} (r_t - b)
\end{align}
where the baseline $b$ is the moving average of $r_t$ within a batch to eliminate reward scaling effects. An entropy regularization term is added to encourage policy exploration and prevent premature collapse:
\begin{align}
\mathcal{L}_{\text{Ent}} = -\eta \sum_a \pi_\theta(a \mid state_t) \log \pi_\theta(a \mid state_t)
\end{align}
The final RABC training objective is a weighted sum of the three terms:
\begin{align}
\mathcal{L}_{\text{RABC}}(\theta) = \beta_1 \mathcal{L}_{\text{BC}} + \beta_2 \mathcal{L}_{\text{PO}} + \mathcal{L}_{\text{Ent}}
\end{align}
where \( \beta_1 \) and \( \beta_2 \) are balancing weights.

\subsubsection{Coordination mechanism with SSDF-Core}
During the inference stage, SSDF-Navigator and SSDF-Core-DPO work synergistically to enable efficient and clinically coherent multi-turn proactive consultation. 

In this process, the system dynamically maintains a fixed-length historical symptom queue $\mathbf{H}_t$ that stores only recent dialogue symptoms mappable to the standard symptom space. At the initial turn ($t=0$), $\mathbf{H}_t$ is empty, and the inquiry $q_0$ generated by SSDF-Core-DPO is adopted directly. For subsequent turns ($t \geq 1$), the synergy proceeded in three key steps: 1) \textbf{Queue update}: $\mathbf{H}_t$ is updated according to the mapping result of $q_{t-1}$. If $q_{t-1}$ can be mapped to a standard symptom $s_{t-1}$, $s_{t-1}$ is appended to $\mathbf{H}_t$, with the earliest symptom removed when the length limit $L$ is exceeded. 2) \textbf{Navigator activation and candidate generation}: SSDF-Navigator is activated only if $q_{t-1}$ is successfully mapped and $\mathbf{H}_t$ is non-empty. Taking $\mathbf{H}_t$ as input, it outputs a probability distribution over the 83 symptoms. The top‑5 symptoms are converted to standardized questions, forming a candidate set $C_t^{\text{nav}}$. Meanwhile, SSDF-Core-DPO generates a candidate inquiry $\tilde{q}_t$ independently based on the dialogue history. 3) \textbf{Candidate-based inquiry selection}: SSDF-Core-DPO selects the most appropriate next question from the combined candidate set $C_t = C_t^{\text{nav}} \cup {\tilde{q}_t}$, guided by the full dialogue history and the clinical logic for spleen‑stomach disorders. The selected question serves as the actual inquiry $q_t$. 

This mechanism integrates the navigator’s global view of symptom progression with the core model’s deep contextual understanding, ensuring each inquiry is strategically guided and clinically reasonable.

Consultation termination is determined primarily by SSDF-Core-DPO, which integrates all collected information to proactively conclude the consultation and generate the final diagnosis and treatment plan. A hard limit of 30 dialogue turns is imposed to avoid overly long interactions, in line with the typical length of clinical spleen-stomach disorder consultations, ensuring practical and efficient dialogues.

The design incorporates three key considerations: conditional activation of SSDF-Navigator to avoid noise under insufficient information; a candidate‑based selection mechanism that leverages both models' strengths for robust decision‑making; and a fixed‑length queue to provide an effective state representation for the navigator while filtering out outdated information. This collaborative design delivers smooth consultation logic, systematic clinical information collection, and faster diagnostic convergence.

\subsection{Construction of SSDF-Bench}
To systematically evaluate models for ICWM-based spleen-stomach diagnosis, we constructed SSDF-Bench, a benchmark covering multidimensional tasks and adopting process-oriented metrics rather than focusing solely on output accuracy.

SSDF-Bench is built on two data sources: 100 de-identified real-world post-recovery clinical cases of spleen-stomach disorders and related test questions extracted from authoritative medical question banks. This composition grounds the benchmark in real clinical practice while aligning it with standardized medical knowledge. Its data preprocessing, expert quality control, and annotation dispute resolution follow the same standardized protocols as our three aforementioned self-constructed datasets. The benchmark achieves an inter-annotator agreement Kappa coefficient of 0.82 and an annotation disagreement rate below 10\%. After strict quality screening and arbitration, the finalized SSDF-Bench comprises 500 high-quality evaluation samples.

SSDF-Bench integrates three evaluation mechanisms to comprehensively assess model outputs and reasoning processes. The first is \textbf{objective evaluation}, including single-choice and multiple-choice questions. For single-choice items, answers are considered correct only if the model produces consistent outputs across three independent reasoning runs and matches the standard answer. For multiple-choice questions, the score is calculated with Eq.\ref{formula:multi-choice} to  the identification of correct options and the exclusion of incorrect ones:
\begin{align} \label{formula:multi-choice}
S = \frac{|A \cap B|}{|A| + |\overline{A} \cap B|}
\end{align}
where $A$ denotes the standard answer set and  $B$ the model’s output set. The second is \textbf{judge model-based subjective evaluation}. We adopt Qwen3-32B as the judge model and use expert-designed structured multi-dimensional scoring prompts to quantitatively assess the rationality of model-generated diagnoses and treatment plans. The third is \textbf{process-oriented evaluation}, which defines two metrics: CoT (Chain of Thought) completeness and CoT accuracy. Using a prompt-based scoring scheme, these two metrics quantify the coverage of key symptoms and examination information in the model’s outputs, as well as the alignment of its reasoning logic with that of clinical physicians.

Based on the above-mentioned evaluation mechanism, SSDF-Bench encompasses six tasks :
\begin{itemize}
    \item \textbf{ICWM Syndrome Differentiation Diagnosis of Spleen-Stomach Disorders (DDSSD)}: one core task, covering 10 dimensions. 
    \item \textbf{Multi-turn Proactive Consultation for ICWM Spleen-Stomach Disorder Diagnosis (MPC)}: the other core task, covering 2 dimensions.
    \item \textbf{Basic knowledge of Traditional Chinese Medicine (TCM-BC)}: covering fundamental TCM theories, etiology and pathogenesis, diagnostic methods, and syndrome differentiation principles, as well as classical herbal formulas and their properties.
    \item \textbf{Basic knowledge of Western Medicine (WM-BC)}: covering human anatomy and physiology, pathological mechanisms of common diseases, diagnostic techniques, and clinical pharmacology of major drug classes.
    \item  \textbf{Medical Security (MS)}: assessing clinical boundary compliance and medical safety value alignment.
    \item \textbf{Medical Ethics (ME)}:evaluating clinical ethics understanding and judgment.
\end{itemize}

\section{Experiments}

All experiments in this study were conducted on an 80 GB A100 GPU. To comprehensively evaluate the performance of SSDF-Core, we selected 12 mainstream LLMs as baseline models (§\ref{sec:baselines}), and conducted experimental analyses on four core aspects: 1) comparative performance on SSDF-Bench (§\ref{sec:SSDF-Bench}); 2) ablation experiments on the two-stage training strategy (§\ref{sec:Two-stage}); 3) validation of the effectiveness of SSDF-Navigator (§\ref{SSDF-Navigator}); 4) validation of the effectiveness of LLM-as-a-Judge used in SSDF-Bench (§\ref{sec:LLM-as-a-Judge}).

\subsection{Baselines}
\label{sec:baselines}
Baseline models cover diverse architectures and parameter scales, all deployed and evaluated under unified settings. Specifically, small open-source models were locally deployed using the vLLM inference framework (v0.15.0) \cite{VLLM2023}, while large open-source models and all closed-source models were evaluated via official APIs. All model calls followed the OpenAI-compatible interface specification, with the temperature parameter fixed at 0 to ensure output stability; quantitative CoT analysis was disabled for models unable to generate complete chain-of-thought content. 

The 12 baseline models fall into three categories. \textbf{General LLMs} include GPT-5.2 \cite{GPT52026}, DeepSeek-V3.2 \cite{liu2024deepseek}, Qwen3-32B \cite{yang2025qwen3}, GLM4.5-9B \cite{glm2024chatglm}, and Gemini-3 \cite{gemini2023multimodal}. \textbf{Medical LLMs} include HuatuoGPT-o1-7B \cite{chen2024huatuogpt}, Baichuan-M3-245B \cite{Dou2025BaichuanM2SM}, and MMed-LM-8B \cite{singhal2023large}. \textbf{TCM-Specific LLMs} include Zhongjing-LLaMa-lora \cite{zhongJing2025}, Sunsimiao-Qwen2-7B \cite{xin2023sunsimiao}, BianCang-14B \cite{Wei2024BianCangAT}, and ShizhenGPT-32B-LLM \cite{Chen2025ShizhenGPTTM}.

\subsection{Performance Comparison on SSDF-Bench}
\label{sec:SSDF-Bench}


\begin{table*}[htbp]
\caption{Performance Comparison Results on SSDF-Bench}
\centering
\renewcommand{\arraystretch}{1.19}
\resizebox{0.9\textwidth}{!}{
\begin{tabular}{l|l|c|c|c|c|c|c}
\toprule
\bf Type & \bf Model & \bf ME & \bf LLMS & \bf TCM-BC & \bf WM-BC & \bf DDSSD & \bf Total score \\
\midrule
\multirow{5}{*}{\bf General LLMs} 
& DeepSeek-V3.2 (w/) & 0.81 & 0.89 & 0.8143 & \textbf{0.8358} & 0.5994 & 0.7398 \\
& Gemini-3 (w/) & 0.51 & 0.70 & 0.8617 & 0.8158 & 0.6010 & 0.6969 \\
& GPT-5.2-2025-12-11 (w/) & \textbf{0.9} & 0.82 & 0.5819 & 0.6794 & 0.6051 & 0.6663 \\
& GLM-4.5 (w/) & 0.77 & 0.96 & 0.7803 & 0.7825 & 0.6063 & 0.7280 \\
& Qwen3-32B (w/) & 0.75 & 0.97 & 0.7651 & 0.7069 & 0.5323 & 0.6793 \\
\midrule
\multirow{4}{*}{\bf TCM-Specific LLMs}
& Biancang-14B (w/o) & 0.88 & 0.97 & 0.7507 & 0.7275 & 0.4542 & 0.6623 \\
& ShizhenGPT-32B-LLM (w/o) & 0.86 & 0.96 & 0.7592 & 0.6733 & 0.5087 & 0.6720 \\
& Sunsimiao-Qwen2-7B (w/o) & 0.75 & 0.89 & 0.7497 & 0.5805 & 0.3102 & 0.5541 \\
& Zhongjing-LLaMA-lora (w/o) & 0.17 & 0.57 & 0.0637 & 0.0498 & 0.0456 & 0.1149 \\
\midrule
\multirow{3}{*}{\bf Medical LLMs}
& Baichuan-M3 (w/) & 0.74 & 0.77 & 0.7768 & 0.7508 & 0.5904 & 0.6927 \\
& HuatuoGPT-o1 (w/) & 0.77 & 0.96 & 0.6417 & 0.6767 & 0.4823 & 0.6296 \\
& MMedLM-8B (w/o) & 0.19 & 0.66 & 0.0945 & 0.0927 & 0.1309 & 0.1748 \\
\midrule
\multirow{1}{*}{\bf OURS}
& SSDF-Core-DPO (w/) & 0.89 & \textbf{0.97} & \textbf{0.8759} & 0.8231 & \textbf{0.7457} & \textbf{0.8241} \\
\midrule
\end{tabular}
}
\label{tab:ssdf-bench-eval}
\end{table*}

We evaluated all aforementioned baselines and our proposed SSDF-Core on SSDF-Bench\footnote{The MPC task is excluded in performance comparison, as baseline models lack the capability for ICWM-oriented multi-turn proactive consultation.}. The overall results are presented in Table \ref{tab:ssdf-bench-eval}\footnote{Models with explicit reasoning abilities are marked as "w/" in the table, otherwise marked as "w/o"}. In terms of overall performance, SSDF-Core-DPO ranks first on SSDF-Bench, significantly outperforming the top models in each baseline category: it surpasses the best general LLM (DeepSeek-V3.2) by \textbf{11.4\%}, the leading TCM-specific LLM (ShizhenGPT-32B-LLM) by \textbf{22.6\%}, and the top medical LLM (Baichuan-M3) by \textbf{19.0\%}. These results validate the effectiveness of our dedicated datasets and tailored training strategy for ICWM spleen-stomach disorder diagnosis.

It is important to note that while most baseline models excel on basic knowledge tasks owing to their rich general and medical knowledge from large-scale pre-training, they perform considerably worse on the core DDSSD task, which demands specialized clinical reasoning and deep integration of ICWM knowledge tailored to spleen-stomach disorder diagnosis. This discrepancy indicates that most models only hold static knowledge reserves but fail to translate such knowledge into effective dynamic clinical reasoning. In contrast, SSDF-Core-DPO significantly outperforms all baselines on the DDSSD task while maintaining top-tier performance on basic knowledge tasks, surpassing the second-place model GLM-4.5 by \textbf{22.99\%} and the third-place model GPT-5.2 by \textbf{23.24\%}. This substantial gap demonstrates that SSDF-Core-DPO effectively integrates general medical knowledge with specialized ICWM diagnostic capabilities for spleen-stomach disorders, which is supported by high-quality ICWM specialty data and the ICWM-specialized training strategy.

\begin{figure*}[htbp]
\centering
\includegraphics[width=0.8\textwidth]{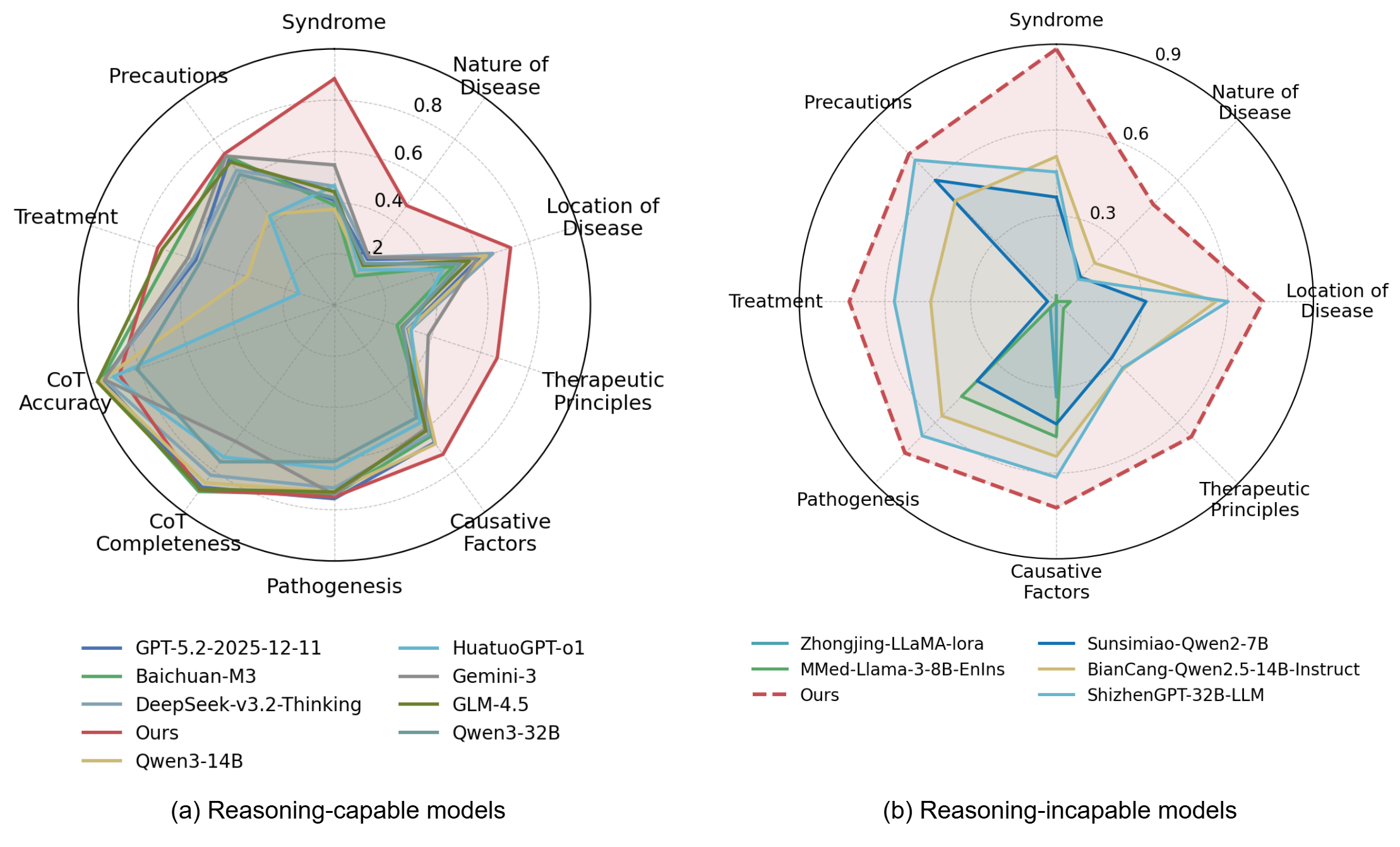}
\caption{Performance Comparison Across Dimensions of the DDSSD Task.}
\label{fig:DDSSD}
\end{figure*}

Figures \ref{fig:DDSSD}(a) and (b) further present the score distributions across dimensions of the DDSSD task for models with and without explicit reasoning abilities, respectively. Overall, reasoning-capable models outperform reasoning-incapable models across all dimensions, verifying that reasoning capacity is a key prerequisite for stable performance in complex clinical syndrome differentiation. Among reasoning-capable models, SSDF-Core-DPO achieves high and more balanced scores across all dimensions, whereas other comparable models show clear weaknesses in certain sub-dimensions. Notably, SSDF-Core-DPO significantly surpasses all baselines in three core dimensions: Syndrome, Nature of Disease, and Therapeutic principles. These dimensions correspond respectively to the core outcomes of syndrome differentiation, the essence of the disease, and treatment direction, representing the most critical links in the clinical decision-making process. The outstanding performance of SSDF-Core-DPO on these dimensions reflects its superior ability to capture syndrome essence and formulate rational therapeutic principles for real-world ICWM diagnostic scenarios of spleen-stomach disorders, thereby providing reliable support for subsequent treatment planning.

\subsection{Effectiveness of Two-stage Training Strategy}
\label{sec:Two-stage}
To systematically evaluate the contribution of each training stage, we perform ablation experiments on three model variants: (i) the base model without further training, (ii) the model trained with SFT only (SSDF-Core w/o DPO), and (iii) the full model with both SFT and DPO (SSDF-Core-DPO).

Table \ref{tab:ssdf-core-ablation} summarizes the results across all tasks on SSDF-Bench. Overall, each training stage yields progressive performance gains. Specifically, the base model yields relatively lower performance due to the lack of task-specific adaptation. After SFT, SSDF-Core w/o DPO achieves substantial improvements on most tasks, validating that task-specific instruction tuning effectively activates the model’s core clinical capabilities. Furthermore, SSDF-Core-DPO consistently outperforms both the base model and the SFT-only variant, with the most significant improvements on the DDSSD task. This confirms that preference alignment further refines the model’s reasoning quality and clinical reliability.

Further analysis of the multi-turn proactive consultation (MPC) task reveals an intriguing and counterintuitive pattern. The base model achieves moderate performance owing to its general conversational ability, yet SSDF-Core w/o DPO exhibits a drop of approximately 18.2\% after SFT. This implies that domain-specific instruction fine-tuning may partially overwrite the general interaction patterns underlying fluent multi-turn dialogue, a typical trade-off when adapting general LLMs to domain-specific tasks via SFT. After DPO, however, SSDF-Core-DPO not only recovers but also outperforms the base model by approximately \textbf{6.4\%},  demonstrating that preference alignment reconciles domain specialization with natural conversational fluency.

Collectively, these results verify the effectiveness of the proposed two-stage training paradigm, in which DPO serves a key role in balancing high-quality clinical reasoning and robust conversational performance.

\begin{table}[htbp]
\caption{Ablation Results of the two-stage training strategy on SSDF-Bench}
\centering
\renewcommand{\arraystretch}{1.19}
\resizebox{0.5\textwidth}{!}{
\begin{tabular}{l|c|c|c|c|c|c}
\toprule
\bf Model & \bf ME & \bf LLMS & \bf TCM-BC & \bf WM-BC & \bf DDSSD & \bf MPC \\
\midrule
\bf Qwen3-14B (base)  & 0.73 & 0.97 & 0.7162 & 0.6847 & 0.5499 & 0.5702  \\
\bf SSDF-Core w/o DPO  &  0.85 & \textbf{0.99} & 0.7861 & \textbf{0.8275} & 0.6995 & 0.4663 \\
\bf SSDF-Core-DPO  & \textbf{0.89} & 0.97 & \textbf{0.8759} & 0.8231 & \textbf{0.7457} & \textbf{0.6066} \\
\midrule
\end{tabular}
}
\label{tab:ssdf-core-ablation}
\end{table}

\subsection{Effectiveness of the Consultation Navigation model}
\label{SSDF-Navigator}
To evaluate the synergistic effect of SSDF‑Navigator on the MPC task, we conducted a controlled ablation study across two configurations: the core diagnostic LLM supported by the consultation navigation model (SSDF-Core w/o Navigator) and SSDF-Core only (SSDF-Core w/ Navigator).

Two dimensions of the MPC task are defined as follows:
\begin{itemize}
    \item \textbf{Critical Symptom Recall}: This metric quantifies the proportion of critical symptoms elicited by the model under evaluation across all test dialogue instances:
    \begin{align}
      \text{Recall}_{\text{crit}} = \frac{|S_{\text{elicited}} \cap S_{\text{gold}}|}{|S_{\text{gold}}|}
    \end{align}
    where $S_{\text{gold}}$ denotes the set of ICWM expert-defined critical symptoms, and $S_{\text{elicited}}$ is the set of symptoms corresponding to the questions elicited by the model.
    \item \textbf{Average Dialogue Efficiency}: This metric measures the average number of clinically relevant symptoms acquired per dialogue turn: 
      \begin{align}
      \text{Eff}_\text{dia} = \frac{N_{\text{valid}}}{N}
      \end{align}
    where $N_{\text{valid}}$  is the total number of symptoms corresponding to the questions elicited by the model that appear in gold-standard case records (including both critical and non-critical symptoms), and $N$ is the total number of dialogue turns across all test dialogue instances.
\end{itemize}

As shown in Table \ref{tab:ablation-navigator}, with the introduction of SSDF-Navigator, SSDF-Core achieved significant improvements in both dimensions. Specifically, $\text{Recall}_{\text{crit}}$ increases by \textbf{7.96} percentage points, indicating that SSDF-Navigator facilitates more systematic coverage of clinically essential inquiry points anligns the inquiry path more closely with the standard clinical diagnostic logic: chief complaint → core symptoms → triggers → accompanying symptoms → syndrome convergence. Meanwhile, $\text{Eff}_{\text{dia}}$ rises by \textbf{7.6} percentage points, demonstrating that SSDF-Navigator’s strategic guidance allows SSDF-Core to collect sufficient diagnostic evidence within a limited number of dialogue turns while reducing redundant or low-information questions.  The simultaneous improvement in both metrics confirms that SSDF-Navigator effectively strengthens the core model’s control over the inquiry process via strategic guidance.

\begin{table}[htbp]
\caption{Ablation results of SSDF-Navigator on the MPC task}
\centering
\renewcommand{\arraystretch}{1.2}
\begin{tabular}{l|c|c}
\toprule
\bf Configuration / Demension & \bf $\text{Recall}_{\text{crit}}$ & \bf $\text{Eff}_\text{dia}$ \\
\midrule
SSDF-Core w/o Navigator &  0.6080 (357/653) & 0.3444 \\
SSDF-Core w/ Navigator & \textbf{0.6876} (449/653) & \textbf{0.3701} \\
\midrule
\end{tabular}
\label{tab:ablation-navigator}
\end{table}

\subsection{Effectiveness of LLM-as-a-Judge}
\label{sec:LLM-as-a-Judge}
To verify the reliability of the LLM-as-a-Judge used in SSDF-Bench for subjective evaluation, we conducted a controlled experiment comparing the scoring behavior of the judge model on semantically consistent answers versus deliberately perturbed answers. 

\begin{figure}[t]
\centering
\includegraphics[width=0.49\textwidth]{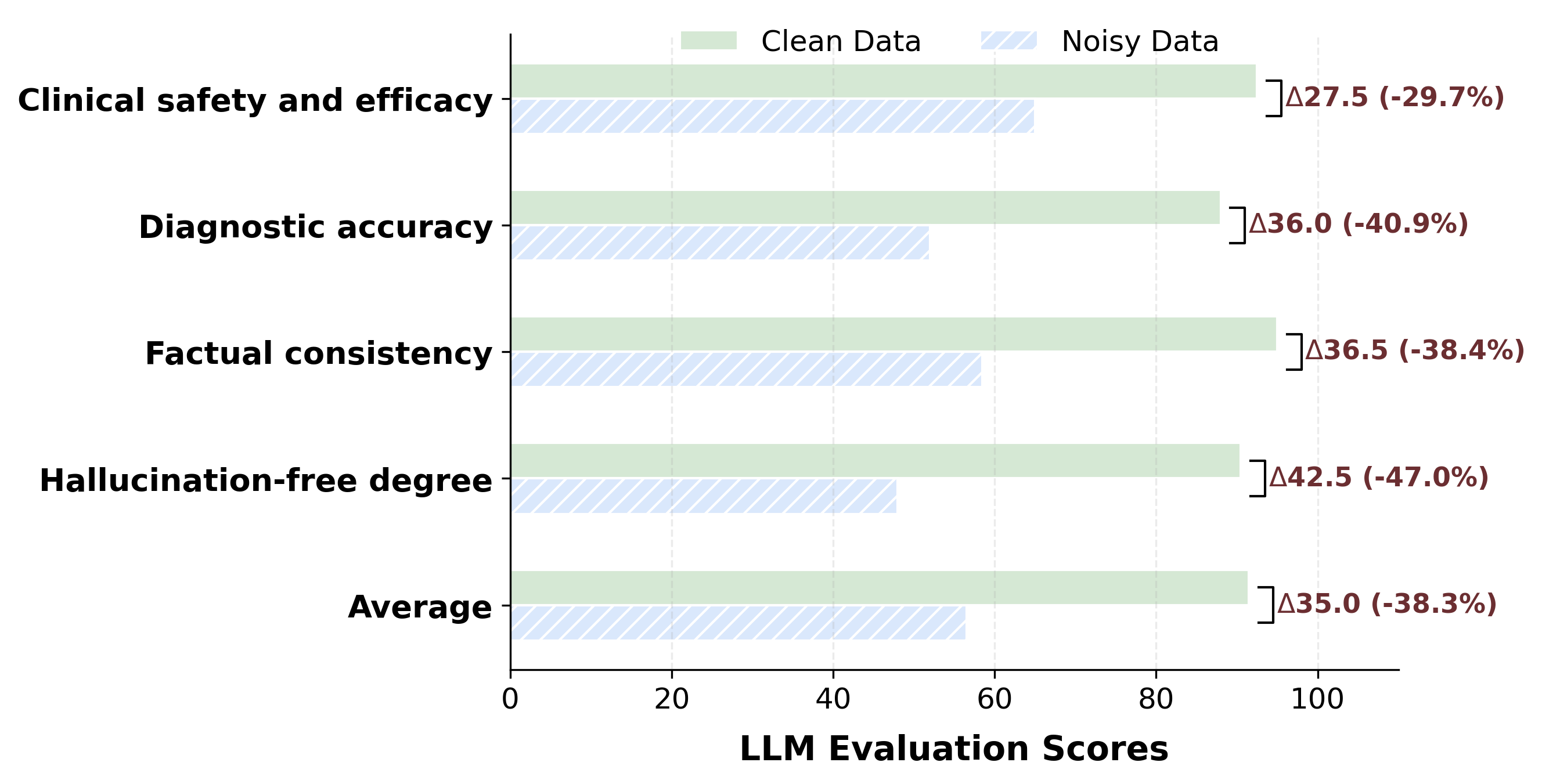}
\caption{Effectiveness Evaluation of the Judge Model in SSDF-Bench.}
\label{fig:LLM-as-a-judge}
\end{figure}

Two datasets were constructed from expert-validated ground-truth responses: 1) \textbf{Clean Data}: paraphrased versions retaining all key entities and causal logic; 2) \textbf{Noisy Data}: fluent but semantically flawed versions with altered critical medical entities or relations. Perturbations were designed across four clinically relevant dimensions (see Fig. \ref{fig:LLM-as-a-judge}).

The judge model was evaluated from three perspectives:
\begin{itemize}
    \item \textbf{Sensitivity}: quantified by the relative score drop between clean and perturbed responses:
    \begin{align}
        \text{S}_{\text{drop}} = \frac{\text{S}_{\text{c}} - \text{S}_{\text{p}}}{\text{S}_{\text{c}}}
    \end{align}
    where $\text{S}_{\text{c}}$ and $\text{S}_{\text{p}}$ are average scores for clean and perturbed responses. A larger drop indicates stronger sensitivity to clinical errors, with a drop above 40\% representing strong error discrimination.
    \item \textbf{Robustness}: evaluated by scoring consistency for semantically equivalent clean rewrites.
   \item \textbf{Expert Alignment}: measured by Spearman correlation between model scores and ICWM expert ratings on 50 randomly sampled responses.
\end{itemize}

As shown in Fig \ref{fig:LLM-as-a-judge}, the judge model demonstrated high stability on clean data and strong sensitivity to perturbations, with an overall score drop of \textbf{38.3\%}. It achieves the highest sensitivity on Hallucination‑Free Degree (drop of \textbf{47.0\%}), proving its ability to detect fabricated medical content. Factual Consistency yields the highest clean scores, confirming the model focuses on semantic fidelity rather than surface phrasing. A clear score reduction (\textbf{29.7\%}) is also observed for Clinical Safety and Efficacy under subtle contraindication perturbations. These results validate that LLM-as-a-Judge serves as a sensitive, robust, and clinically consistent evaluator for ICWM diagnostic reasoning tasks.

\section{Conclusion}
This study targets the core challenges in intelligent diagnosis of spleen-stomach disorders under the Integrative Chinese and Western Medicine (ICWM) paradigm, and proposes DongYuan, a  unified framework that integrates specialized datasets, domain-adapted models, and a standardized evaluation benchmark. At the data level, we constructed and open-sourced three ICWM datasets, laying a high-quality, standardized foundation for domain-specific model training. At the model level, we introduced SSDF-Core, a core diagnostic LLM trained via a two-stage regimen for deep domain adaptation, together with SSDF-Navigator, a pluggable consultation navigation model that enhances the strategicity of multi-turn proactive consultation through hybrid training. At the evaluation level, we developed SSDF-Bench, a comprehensive benchmark centering on assessing model performance in ICWM spleen-stomach disorder diagnosis. Experiments demonstrate that SSDF-Core significantly outperforms mainstream LLMs on SSDF-Bench. Collectively, DongYuan offers a systematic methodology for developing specialized diagnostic artificial intelligence in ICWM, providing a scalable blueprint for extending similar approaches to other disease domains within the ICWM context.

Despite these advances, limitations remain. Methodologically, SSDF-Core currently relies solely on text input and does not incorporate multimodal data that are valuable in ICWM diagnosis. In practical deployment, its computational overhead and interoperability with hospital information systems pose implementation challenges. Future work will explore multimodal fusion within the ICWM paradigm, and focus on model lightweighting, as well as standardized deployment interfaces to facilitate clinical adoption.

\section*{Acknowledgment}
This study was supported by the Scientific and Technological Innovation Project of China Academy of Chinese Medical Sciences (No.ZN2023A02), the "AI+ Health Collaborative Innovation Cultivation Reserve Project" of Beijing Municipal Science and Technology Commission (No.Z251100006025024).

\section*{References}
\bibliographystyle{ieeetr}
\bibliography{ref}
\end{CJK}
\end{document}